\documentclass{article}
\usepackage{spconf,amsmath,graphicx,hyperref}

\usepackage{graphicx}
\usepackage{booktabs}
\usepackage{amssymb}
\usepackage[misc]{ifsym}

\usepackage{multirow}
\usepackage{multicol}

\title{Revisiting the Seasonal Trend Decomposition for Enhanced Time Series Forecasting}
\name{Sanjeev Panta$^{\star}$ \qquad Xu Yuan$^{\dagger}$ \qquad Li Chen$^{\star}$  \qquad Nian-Feng Tzeng$^{\star}$
\thanks{The research was supported by the NSF under OIA-2327452, OIA-2019511, 2425812, 2348452, and by the BoRSF under LEQSF(2024-27)-RD-B-03.}
}
\address{$^{\star}$ University of Louisiana at Lafayette, Lafayette, LA, USA \\
    $^{\dagger}$ University of Delaware, Newark, DE, USA}
\begin{document}
%
\maketitle
\begin{abstract}
Time series forecasting presents significant challenges in real-world applications across various domains. Building upon the decomposition of the time series, we enhance the architecture of machine learning models for better multivariate time series forecasting. To achieve this, we focus on the trend and seasonal components individually and investigate solutions to predict them with less errors. Recognizing that reversible instance normalization is effective only for the trend component, we take a different approach with the seasonal component by directly applying backbone models without any normalization or scaling procedures. Through these strategies, we successfully reduce error values of the existing state-of-the-art models and finally introduce dual-MLP models as more computationally efficient solutions. Furthermore, our approach consistently yields positive results with around 10\% MSE average reduction across four state-of-the-art baselines on the benchmark datasets. We also evaluate our approach on a hydrological dataset extracted from the United States Geological Survey (USGS) river stations, where our models achieve significant improvements while maintaining linear time complexity, demonstrating real-world effectiveness. The source code is available at \href{https://github.com/Sanjeev97/Time-Series-Decomposition}{https://github.com/Sanjeev97/Time-Series-Decomposition}
\end{abstract}
\section{Introduction}
\label{sec:intro}


The forecasting of time series data has substantial practical implications across domains including weather forecasting, electricity consumption estimation, and streamflow forecasting. Existing methods range from traditional statistical tools to advanced techniques such as Recurrent Neural Networks (RNNs), Long Short Term Memory networks (LSTMs), Convolutional Neural Networks (CNNs), and Temporal Convolution Networks (TCNs), but fall short in computation parallelism and long-range dependency capture compared to attention mechanisms \cite{wen2022transformers}.

The advent of the Transformer architecture marked a significant revolution in Natural Language Processing (NLP) \cite{vaswani2017attention} as well as various other domains, thanks to the powerful self-attention mechanism. 
However, canonical self-attention is poorly suited for time series due to prohibitive computational complexity for long sequences. Existing efforts \cite{zhou2021informer,zhou2022fedformer,zhang2022first} adapt self-attention to sparse versions, trading information utilization for efficiency. Autoformer \cite{wu2021autoformer} replaced self-attention with auto-correlation and integrated time series decomposition \cite{anderson1976time,cleveland1990stl} to break series into trend-cyclic and seasonal components, enabling series-wise representation aggregation. Such decomposition-enhanced time series forecasting  has inspired numerous Transformer-based models \cite{ouyang2023rankformer,ouyang2023stlformer,zhou2022fedformer,woo2022etsformer,zhang2022first} integrating decomposition mechanisms, as a crucial part. They mostly incorporate Mixture of Experts \cite{zhou2022fedformer} or Series Decomposition \cite{cleveland1990stl} in their adapted manners, with rather sophisticated architectures, including layers of encoder-decoder blocks with unnecessarily complex decomposition \cite{woo2022etsformer,zhou2022film}. 
In this paper, we are motivated to make a simpler design by relying on the basic moving average decomposition \cite{anderson1976time}.

Classical decomposition methods include moving average and alternatives like STL \cite{cleveland1990stl}, VMD \cite{dragomiretskiy2013variational}, RobustSTL \cite{wen2019robuststl}, OnlineSTL \cite{mishra2021onlinestl}, and EEMD \cite{mhamdi2011trend}. Despite advanced techniques, moving average remains attractive and widely used. Recent architectures include Rankformer \cite{ouyang2023rankformer} with Multi-Level Decomposition, STLformer \cite{ouyang2023stlformer} with STL decomposition, ETSformer \cite{woo2022etsformer} with exponential smoothing attention, and FEDformer \cite{zhou2022fedformer} with frequency-enhanced blocks. Linear architectures \cite{zeng2023transformers} challenge complexity assumptions, while iTransformer \cite{liu2023itransformer} uses inverted dimensions and PatchTST \cite{nie2022time} employs patching mechanisms.

\begin{figure*}[!thbp]
\centering
\includegraphics[height=6.2cm, width=0.8\textwidth]{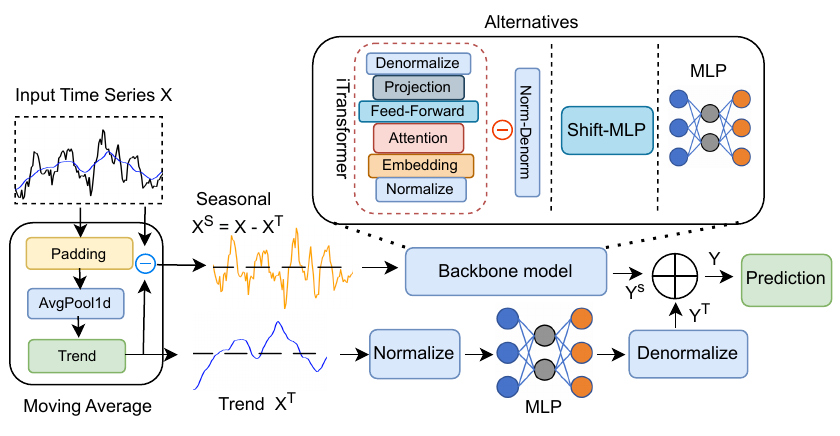}
\caption{Illustration of our approach with moving average and showing iTransformer (as backbone) with two other alternatives: Shift-MLP and MLP.} \label{fig:approach}
\end{figure*}

Unlike them, our proposed design breaks the original time series into trend and seasonal components, each of which undergoes training and prediction separately. Eventually, they are combined again to form the predicted outcome. To avoid the difficulty in the selection of normalization and scaling for the seasonality component \cite{ye2024frequency}, we take a different approach. We eliminate the need for these techniques by removing normalization from state-of-the-art (SOTA) models and adapting them to serve as backbones for seasonal processing, which significantly increases their accuracy. Additionally, we introduce two new dual-MLP models that replace these SOTA backbones with Multilayer Perceptrons (MLP), both achieving better results and greater efficiency than existing approaches.

Our contributions include leveraging existing SOTA architectures as backbones, and systematically improving their forecasting capabilities, by adapting moving average decomposition. Our approach has been tested with various benchmark datasets for multivariate time series forecasting and models including iTransformer \cite{liu2023itransformer}, PatchTST \cite{nie2022time}, TimesNet \cite{wu2022timesnet}, and DLinear \cite{zeng2023transformers}, achieving $\sim10\%$ average MSE reduction. Additionally, we propose streamlined dual-MLP architectures that outperform complex counterparts with linear time complexity and superior computational efficiency, reducing MSE by $\sim7\%$ on USGS\footnote{https://waterdata.usgs.gov/nwis/rt} hydrological data from Comite River streamflow gauges in Louisiana. We demonstrate consistent performance improvements across multiple domains while requiring significantly lower computational resources, and extend evaluation to real-world environmental applications, highlighting our method's practical utility in hydrological forecasting.

\section{Methodology}
\label{sec:method}
\subsection{Problem Definition and Approach}
Given multivariate time series input $X \in \mathbb{R}^{L \times C}$, where $L$ is the input series length and $C$ is the number of channels, we predict output $Y \in \mathbb{R}^{H \times C}$ for horizon $H$.

Current time series forecasting faces challenges in prediction accuracy and computational efficiency, with models suffering from slow training times and high costs. This is particularly evident when processing large datasets like USGS hydrological measurements. Existing methodologies have technical constraints: RevIN shows limitations in handling seasonal components, and insufficient exploration of decomposition techniques limits SOTA model performance.

Our framework (Figure \ref{fig:approach}) can be applied to existing time series models for enhancement through strategic decomposition techniques, preferably the simplest moving average decomposition, while we also explore two alternative decomposition methods. Our approach separates input signals into trend and seasonal components, processes them through specialized networks, and recombines them for the final prediction.

\subsection{Decomposition Techniques} 
We investigate three decomposition methods for possible incorporation into our framework:

\textbf{Moving Average Decomposition} follows the classical decomposition method based on moving averages. We decompose the input series into trend and seasonal components. We adopt AvgPool(.) for moving average to smooth the time series, with padding operation to keep the series length unchanged \cite{wu2021autoformer}. We filter out the Trend $X^T$ and then subtract it from the input $X$ to get the Seasonal signal $X^S$. 
\begin{align}\label{eq:AvgPool}
X^T &= AvgPool(Padding(X)) \\
X^S &= X - X^T
\end{align}
\textbf{Mixture of Experts (MoE) Decomposition} \cite{zhou2022fedformer} employs multiple average filters with different sizes to extract multi-scale trend components.
\begin{equation}
\label{eq:MoE}
X_{M}^{T} = Softmax(L(x)) * (F(x))
\end{equation}
Here, ${X_{M}^{T}}$ is the extracted trend, and $F(.)$ is a set of average pooling filters defined by equation \ref{eq:AvgPool} and $Softmax(L(x))$ is the weights for mixing these extracted trends.

\textbf{Frequency-Based Decomposition} separates components using Fourier analysis \cite{zhou2022fedformer,woo2022etsformer}. We extract the seasonal part $X^S$ from the input. We achieve the trend component $X^T$ by removing seasonal component $X^S$ from the input signal $X$.
\begin{align}
    Z &= FT(X) \\
    K &= TopK(Amp(Z)) \\
    X^S &= IFT(Filter(K, Z)) \\
    X^T &= X - X^S
\end{align}
 The input transforms to frequency domain using Fast Fourier Transform (FFT), selects top-$K$ amplitude components using a binary mask filter for seasonal extraction, then reconstructs using inverse FFT. This approach has $O(L\log L)$ complexity due to FFT operations, while the moving average technique achieves $O(L)$, making it the simplest and fastest.

\subsection{Trend Component}
We incorporate Reversible Instance Normalization (RevIN) and Multilayer Perceptrons (MLP) to predict the trend part of the input signal. The fluctuation of mean, variance and seasonality over time, coined non-stationarity in time series, poses challenges for analysis and forecasting. We use RevIN as it is preferred for the trend component \cite{zhang2022first,kim2021reversible}, to remove the non-stationary characteristics and also to restore them after passing the trend component through a three-layered MLP Neural Network. 
\begin{equation}
Y^T = RevIN(MLP(RevIN(X^T)))
\end{equation}
Here, $Y^T$ denotes the predicted trend component obtained and $X^T$ is the initialized trend component.

\subsection{Backbone Model}
The seasonal component, including residual noise, is processed through a backbone network ranging from simple linear models to complex transformer architectures. We remove normalization (shown in Figure \ref{fig:approach}) from existing models such as iTransformer when using them as backbones. This approach preserves seasonal component structure and brings a performance improvement over the original model in terms of reducing forecast errors, as evidenced in our evaluation (Section \ref{sec:results}). Beyond enhancing existing models of sophisticated architectures, we also develop lightweight backbone alternatives (Section \ref{sec:eff-models}) for high computational efficiency while maintaining strong performance. The backbone's output, the predicted seasonal component $Y^S$, is formally expressed as:
\begin{equation}
    Y^S = Backbone(X^S)
\end{equation}
\subsection{Efficient Models}
\label{sec:eff-models}
We introduce two dual-MLP models based on RevIN, decomposition, and MLPs. Since RevIN cannot effectively handle seasonality components, both RMSM and RMM process seasonal components without normalization or scaling, replacing complex backbones with more efficient approaches. Both models achieve linear $O(L)$ complexity,
providing computationally efficient alternatives to complex architectures.

    \textbf{RevIN-MLP-Shift-MLP (RMSM)} processes seasonal components through a three-stage architecture:
\begin{align} \label{eq:relu1}
    h_1 &= ReLU(W_1, X^{S}) \\
    \label{eq:concat}
    h_2 &= ReLU(W_2Concat(h1, X)) \\
    Y^{S} &= ReLU(W_3, h_2)
\end{align}
It captures the seasonal components through a linear transformation with ReLU activation in $h_1$. Then, it concatenates the processed features with the original input to provide context in $h_2$, before generating predictions in $Y^S$. This approach, known as Shift Forecasting \cite{ye2024frequency}, effectively handles the shift in non-stationary information that may evolve between input and output signals.

\textbf{RevIN-MLP-MLP (RMM)} employs a standard 3-layered MLP for seasonal component processing, the same one used for the trend part.
\begin{equation}
    Y^{S} = W_3{ReLU}(W_2{ReLU}(W_1 X^{S}))
\end{equation}
This straightforward approach maintains competitive forecasting accuracy while being simpler to train. Unlike RMSM, RMM processes seasonal components directly without additional context from the original signal.
Finally, the trend prediction $Y^T$ and seasonal prediction $Y^S$ are combined to get the final forecasting result.
\begin{equation}
    Y = Y^T + Y^S
\end{equation}
\section{Experiments}
\label{sec:experiments}
\begin{table*}[!ht]
\caption{Complete performance comparison. Bold: best overall, Underlined: improvements over baselines.}
\centering
\resizebox{0.95\textwidth}{!}{
\begin{tabular}{|c|c|cc|cc|cccc|cccc|cccc|cccc|}
\hline
\multirow{3}{*}{\rotatebox{90}{Dataset}} & & \multicolumn{2}{c|}{RMSM} & \multicolumn{2}{c|}{RMM} & \multicolumn{4}{c|}{iTransformer} & \multicolumn{4}{c|}{PatchTST} & \multicolumn{4}{c|}{TimesNet} & \multicolumn{4}{c|}{DLinear} \\
\cline{3-22}
& & & & & & \multicolumn{2}{c|}{Original} & \multicolumn{2}{c|}{+Ours} & \multicolumn{2}{c|}{Original} & \multicolumn{2}{c|}{+Ours} & \multicolumn{2}{c|}{Original} & \multicolumn{2}{c|}{+Ours} & \multicolumn{2}{c|}{Original} & \multicolumn{2}{c|}{+Ours} \\
\cline{3-22}
& H & MSE & MAE & MSE & MAE & MSE & MAE & MSE & MAE & MSE & MAE & MSE & MAE & MSE & MAE & MSE & MAE & MSE & MAE & MSE & MAE \\
\hline
\multirow{4}{*}{\rotatebox{90}{Weather}} 
& 96 & \textbf{0.159} & \textbf{0.197} & 0.163 & 0.199 & 0.174 & 0.214 & \underline{0.155} & \underline{0.192} & 0.186 & 0.227 & \underline{0.162} & \underline{0.195} & 0.172 & 0.220 & \underline{0.165} & \underline{0.209} & 0.195 & 0.252 & \underline{0.163} & \underline{0.199} \\
& 192 & \textbf{0.200} & \textbf{0.238} & 0.208 & 0.241 & 0.221 & 0.254 & \underline{0.208} & \underline{0.240} & 0.234 & 0.265 & \underline{0.208} & \underline{0.241} & 0.219 & 0.261 & \underline{0.211} & \underline{0.247} & 0.237 & 0.295 & \underline{0.207} & \underline{0.240} \\
& 336 & \textbf{0.248} & \textbf{0.276} & 0.263 & 0.282 & 0.278 & 0.296 & \underline{0.263} & \underline{0.281} & 0.284 & 0.301 & \underline{0.262} & \underline{0.281} & 0.246 & 0.337 & 0.270 & \underline{0.289} & 0.282 & 0.331 & \underline{0.262} & \underline{0.281} \\
& 720 & \textbf{0.323} & \textbf{0.330} & 0.342 & 0.334 & 0.358 & 0.347 & \underline{0.340} & \underline{0.333} & 0.356 & 0.349 & \underline{0.341} & \underline{0.334} & 0.365 & 0.359 & \underline{0.350} & \underline{0.341} & 0.345 & 0.382 & \underline{0.342} & \underline{0.334} \\
\hline
\multirow{4}{*}{\rotatebox{90}{Electricity}} 
& 96 & 0.151 & 0.236 & 0.155 & 0.239 & \textbf{0.148} & 0.240 & \underline{0.136} & \underline{0.226} & 0.190 & 0.296 & \underline{0.152} & \underline{0.236} & 0.168 & 0.272 & \underline{0.142} & \underline{0.241} & 0.210 & 0.302 & \underline{0.152} & \underline{0.236} \\
& 192 & 0.163 & \textbf{0.247} & 0.165 & 0.249 & \textbf{0.162} & 0.253 & \underline{0.159} & \underline{0.249} & 0.199 & 0.304 & \underline{0.164} & \underline{0.247} & 0.184 & 0.322 & \underline{0.156} & \underline{0.254} & 0.210 & 0.305 & \underline{0.164} & \underline{0.247} \\
& 336 & \textbf{0.178} & \textbf{0.264} & 0.180 & 0.265 & \textbf{0.178} & 0.269 & 0.178 & 0.271 & 0.217 & 0.319 & \underline{0.179} & \underline{0.263} & 0.198 & 0.300 & \underline{0.169} & \underline{0.269} & 0.223 & 0.319 & \underline{0.179} & \underline{0.263} \\
& 720 & 0.218 & 0.311 & \textbf{0.216} & \textbf{0.296} & 0.225 & 0.317 & \underline{0.216} & \underline{0.304} & 0.258 & 0.352 & \underline{0.216} & \underline{0.295} & 0.220 & 0.320 & \underline{0.216} & \underline{0.295} & 0.258 & 0.350 & \underline{0.216} & \underline{0.295} \\
\hline
\multirow{4}{*}{\rotatebox{90}{ETTm2}} 
& 96 & 0.176 & 0.255 & \textbf{0.172} & \textbf{0.249} & 0.180 & 0.264 & \underline{0.167} & \underline{0.246} & 0.183 & 0.270 & \underline{0.172} & \underline{0.250} & 0.187 & 0.267 & \underline{0.172} & \underline{0.250} & 0.193 & 0.293 & \underline{0.172} & \underline{0.249} \\
& 192 & 0.257 & 0.312 & \textbf{0.236} & \textbf{0.292} & 0.250 & 0.309 & \underline{0.231} & \underline{0.289} & 0.255 & 0.314 & \underline{0.236} & \underline{0.293} & 0.249 & 0.309 & \underline{0.236} & \underline{0.293} & 0.284 & 0.361 & \underline{0.236} & \underline{0.292} \\
& 336 & 0.321 & 0.356 & \textbf{0.294} & \textbf{0.330} & 0.311 & 0.348 & \underline{0.288} & \underline{0.327} & 0.309 & 0.347 & \underline{0.295} & \underline{0.331} & 0.321 & 0.351 & \underline{0.290} & \underline{0.334} & 0.382 & 0.429 & \underline{0.294} & \underline{0.330} \\
& 720 & 0.443 & 0.434 & \textbf{0.389} & \textbf{0.387} & 0.412 & 0.407 & \underline{0.385} & \underline{0.385} & 0.412 & 0.404 & \underline{0.389} & \underline{0.388} & 0.408 & 0.403 & \underline{0.396} & \underline{0.395} & 0.558 & 0.525 & \underline{0.389} & \underline{0.387} \\
\hline
\multirow{4}{*}{\rotatebox{90}{ETTh2}} 
& 96 & 0.292 & 0.341 & \textbf{0.276} & \textbf{0.325} & 0.297 & 0.349 & \underline{0.285} & \underline{0.330} & 0.308 & 0.355 & \underline{0.279} & \underline{0.329} & 0.340 & 0.374 & \underline{0.278} & \underline{0.327} & 0.340 & 0.394 & \underline{0.279} & \underline{0.328} \\
& 192 & 0.369 & 0.395 & \textbf{0.346} & \textbf{0.371} & 0.380 & 0.400 & \underline{0.365} & \underline{0.381} & 0.393 & 0.405 & \underline{0.351} & \underline{0.376} & 0.402 & 0.414 & \underline{0.348} & \underline{0.372} & 0.482 & 0.479 & \underline{0.351} & \underline{0.375} \\
& 336 & 0.404 & 0.429 & \textbf{0.356} & \textbf{0.386} & 0.428 & 0.432 & \underline{0.411} & \underline{0.418} & 0.427 & 0.436 & \underline{0.364} & \underline{0.395} & 0.452 & 0.452 & \underline{0.358} & 0.390 & 0.591 & 0.541 & \underline{0.362} & \underline{0.391} \\
& 720 & 0.402 & 0.423 & \textbf{0.401} & \textbf{0.421} & 0.427 & 0.445 & 0.429 & \underline{0.437} & 0.436 & 0.450 & \underline{0.410} & \underline{0.431} & 0.462 & 0.468 & \underline{0.402} & \underline{0.423} & 0.839 & 0.661 & \underline{0.408} & \underline{0.426} \\
\hline
\multirow{4}{*}{\rotatebox{90}{Hydro}} 
& 96 & \textbf{0.112} & \textbf{0.107} & 0.114 & 0.109 & 0.126 & 0.136 & \underline{0.117} & \underline{0.113} & 0.122 & 0.133 & \underline{0.117} & \underline{0.113} & 0.115 & 0.117 & \underline{0.106} & \underline{0.110} & 0.122 & 0.146 & \underline{0.114} & \underline{0.110} \\
& 192 & \textbf{0.237} & \textbf{0.179} & 0.246 & 0.188 & 0.253 & 0.213 & \underline{0.248} & \underline{0.192} & 0.239 & 0.199 & 0.248 & \underline{0.191} & 0.251 & 0.225 & \underline{0.250} & \underline{0.195} & 0.254 & 0.208 & \underline{0.246} & \underline{0.188} \\
& 336 & \textbf{0.356} & \textbf{0.245} & 0.373 & 0.261 & 0.376 & 0.279 & 0.376 & \underline{0.265} & 0.352 & 0.273 & 0.376 & \underline{0.264} & 0.360 & 0.273 & 0.373 & \underline{0.264} & 0.372 & 0.277 & 0.373 & \underline{0.262} \\
& 720 & 0.485 & \textbf{0.318} & 0.519 & 0.345 & 0.501 & 0.338 & 0.525 & 0.349 & 0.478 & 0.329 & 0.523 & 0.348 & 0.530 & 0.367 & \underline{0.520} & \underline{0.347} & \textbf{0.269} & 0.366 & 0.334 & 0.405 \\
\hline
\end{tabular}
}
\label{table:complete_results}
\end{table*}

\begin{table}[!h]
\caption{Ablation Studies. Bold: best, Underlined: second}
\resizebox{\columnwidth}{!}{
\begin{tabular}{|c@{\hspace{4pt}}c|cc|cc|cc||cc|cc|cc|}
\hline
\multirow{3}{*}{\rotatebox{90}{}} & \multirow{3}{*}{H} & \multicolumn{6}{c||}{Backbone} & \multicolumn{6}{c|}{RMM} \\
\cline{3-14}
& & \multicolumn{2}{c|}{MA} & \multicolumn{2}{c|}{MOE} & \multicolumn{2}{c||}{FD} & \multicolumn{2}{c|}{MA} & \multicolumn{2}{c|}{MOE} & \multicolumn{2}{c|}{FD} \\
\cline{3-14}
& & MSE & MAE & MSE & MAE & MSE & MAE & MSE & MAE & MSE & MAE & MSE & MAE \\
\hline
\multirow{4}{*}{\rotatebox{90}{Weather}} & 96 & \textbf{0.155} & \textbf{0.192} & \underline{0.163} & \underline{0.199} & \textbf{0.155} & \underline{0.194} & \underline{0.163} & \underline{0.199} & \textbf{0.162} & \textbf{0.198} & 0.165 & 0.202 \\
& 192 & \underline{0.208} & \textbf{0.240} & \underline{0.208} & \textbf{0.240} & \textbf{0.203} & \underline{0.241} & \underline{0.208} & \underline{0.241} & \underline{0.208} & \textbf{0.240} & \textbf{0.206} & 0.242 \\
& 336 & \textbf{0.263} & \textbf{0.281} & \textbf{0.263} & \textbf{0.281} & \underline{0.265} & \underline{0.291} & \underline{0.263} & \underline{0.282} & 0.264 & \underline{0.282} & \textbf{0.256} & \textbf{0.281} \\
& 720 & \underline{0.340} & \textbf{0.333} & \textbf{0.339} & \textbf{0.333} & 0.356 & \underline{0.349} & 0.342 & 0.334 & \underline{0.341} & \underline{0.333} & \textbf{0.325} & \textbf{0.332} \\
\hline
\multirow{4}{*}{\rotatebox{90}{Hydro}} & 96 & \textbf{0.117} & \underline{0.113} & \textbf{0.117} & \textbf{0.112} & \underline{0.132} & 0.147 & \underline{0.114} & \underline{0.109} & \textbf{0.113} & \textbf{0.108} & 0.117 & 0.111 \\
& 192 & \textbf{0.248} & \underline{0.192} & \underline{0.249} & \textbf{0.190} & 0.257 & 0.216 & 0.246 & 0.188 & \textbf{0.240} & \textbf{0.182} & \underline{0.245} & \underline{0.187} \\
& 336 & 0.376 & \underline{0.265} & \underline{0.375} & \textbf{0.262} & \textbf{0.374} & 0.275 & 0.373 & \underline{0.261} & \textbf{0.363} & \textbf{0.252} & \underline{0.366} & \textbf{0.252} \\
& 720 & 0.525 & \underline{0.349} & \underline{0.515} & \textbf{0.344} & \textbf{0.511} & \textbf{0.344} & 0.519 & 0.345 & \underline{0.496} & \underline{0.330} & \textbf{0.488} & \textbf{0.322} \\
\hline
\multirow{4}{*}{\rotatebox{90}{ETTh2}} & 96 & \textbf{0.285} & \textbf{0.330} & \underline{0.286} & \textbf{0.330} & 0.385 & \underline{0.430} & \textbf{0.276} & \underline{0.325} & \underline{0.277} & \underline{0.325} & 0.289 & \textbf{0.289} \\
& 192 & \underline{0.365} & \underline{0.381} & \textbf{0.360} & \textbf{0.379} & 0.793 & 0.663 & \textbf{0.346} & \underline{0.371} & \textbf{0.346} & \textbf{0.370} & \underline{0.374} & 0.388 \\
& 336 & \underline{0.411} & \underline{0.418} & \textbf{0.406} & \textbf{0.415} & 0.667 & 0.590 & \underline{0.356} & \underline{0.386} & \textbf{0.355} & \textbf{0.385} & 0.381 & 0.405 \\
& 720 & \underline{0.429} & \underline{0.437} & \textbf{0.424} & \textbf{0.435} & 1.120 & 0.807 & \underline{0.401} & \underline{0.421} & \textbf{0.399} & \textbf{0.420} & 0.542 & 0.504 \\
\hline
\end{tabular}
}
\label{table:ablation}
\end{table}

\subsection{Experimental Setup}
We evaluate on four benchmark datasets: Electricity\cite{zhou2021informer}
(321 clients, hourly consumption 2012-2014), ETT \cite{zhou2021informer} (ETTm2 with 15-min intervals, ETTh2 with hourly intervals from transformers 2016-2018), Weather \cite{wu2021autoformer}
(21 meteorological indicators, 10-min intervals, 2020), and Hydrological (4 USGS river stations on Comite River, Louisiana, 15-min gauge heights 2015-2023). All datasets use 7:2:1 chronological train/validation/test splits. We test prediction horizons $H \in \{96, 192, 336, 720\}$ with fixed input length $L = 96$, evaluating using MSE and MAE metrics. 

Our model-agnostic approach applies to any backbone by removing instance normalization and scaling from their implementation. We test iTransformer \cite{liu2023itransformer}, PatchTST \cite{nie2022time}, TimesNet \cite{wu2022timesnet}, and DLinear \cite{zeng2023transformers} in multivariate settings. Implementation uses PyTorch with Adam optimizer (lr=$1 \times 10^{-4}$, batch size=32) on RTX 3080Ti. Model dimension is 512, Shift-MLP uses hidden dimensions [64, 128, 128], MSE serves as training loss, and all experiments repeat three times for averaged results.

\subsection{Results \& Efficiency Analysis}
\label{sec:results}
Our approach demonstrates improved performance across benchmark and hydrological datasets (Table \ref{table:complete_results}). We achieve MSE reductions of 6.81\%, 13.62\%, 9.95\%, and 15.82\% for iTransformer, PatchTST, TimesNet, and DLinear respectively, with MAE reductions of 9.1\%, 12.39\%, 9.35\%, and 11.94\% respectively, across the benchmark datasets. Overall improvements average 9.46\% MSE and 8.89\% MAE for these datasets, with 6.38\% MAE reduction on hydrological data. Our dual-MLP models RMSM and RMM achieve 6.49\% (13.85\%) and 2.98\% (8.99\%) MSE (MAE) improvements over the best SOTA iTransformer while offering superior computational efficiency, preferably for heterogeneous environmental multi-source and homogeneous single-source datasets respectively.
Table \ref{table:complete_results} further shows these lightweight models consistently achieve top performance, demonstrating that simple MLP architectures without normalization can outperform complex models.

Our dual-MLP approaches (RMSM and RMM) achieve $O(L)$ time complexity compared to $O(L^2)$ for Transformer-based models, where $L$ is input window length. Training and inference times are comparable to DLinear (10ms vs 9ms training, 5ms vs 4ms inference) but significantly faster than iTransformer (20ms, 12ms) and TimesNet (136ms training, 75ms inference). Our models provide the second-fastest performance among SOTA methods while achieving superior forecasting accuracy, particularly on weather and hydrological datasets. All timing measurements averaged over 10 runs with $L=96$ and $H=720$ on the hydrological dataset.

\subsection{Ablation Studies}
We evaluate whether computationally expensive backbone models are necessary for superior results. Our RMM model outperforms or matches backbone-enhanced versions (iTransformer) while achieving lower complexity and faster execution. Moving Average (MA) decomposition performs best in most cases, with Frequency Decomposition (FD) and Mixture of Experts (MOE) showing marginal improvements only on Weather and Hydrological datasets as they are known to better handle seasonality extraction \cite{woo2022etsformer}. However, overall best results still come from lightweight MA decomposition with our MLP implementations.
Comparing decomposition methods, Table \ref{table:ablation} shows most optimal values derive from MA and MOE decompositions, with MOE only slightly superior to MA. Results across all three decomposition techniques demonstrate that complex decomposition methods provide minimal improvement over simple moving average, validating our approach's emphasis on simplicity and efficiency.

\section{Conclusion}
We revisited decomposition for time series prediction, finding that simple decomposition adequately improves existing model performance. Expensive backbone or SOTA models aren't necessary for better forecasting i.e. simple MLPs suffice for seasonality without normalization. While Frequency Decomposition and Mixture of Experts offer some improvements, Moving Average remains sufficient.

\bibliographystyle{IEEEbib}
\bibliography{strings,refs}

\end{document}